\pdfoutput=1

\documentclass[11pt]{article}

\usepackage{EMNLP2022}

\usepackage{times}
\usepackage{latexsym}
\usepackage{amsmath}
\usepackage{amssymb}
\usepackage{mathtools}
\usepackage{amsthm}
\usepackage{multicol}
\usepackage{multirow}

\usepackage{graphicx}
\usepackage{subfigure}
\usepackage{booktabs}
\newcommand{\thickhline}{\noalign{\hrule height 1pt}}

\usepackage[T1]{fontenc}

\usepackage[utf8]{inputenc}

\usepackage{microtype}

\usepackage{inconsolata}

\title{Towards Efficient Dialogue Pre-training with Transferable and Interpretable Latent Structure}

\author{
    Xueliang Zhao\textsuperscript{1,3}\footnotemark[1], 
    Lemao Liu\textsuperscript{2}, 
    Tingchen Fu\textsuperscript{4}, 
    Shuming Shi\textsuperscript{2},
    Dongyan Zhao\textsuperscript{1,3,5}\footnotemark[2],
    Rui Yan\textsuperscript{4}\footnotemark[2] \\
    \textsuperscript{1}Wangxuan Institute of Computer Technology, Peking University\\
    \textsuperscript{2}Tencent AI Lab \quad
    \textsuperscript{3}Center for Data Science, AAIS, Peking University\\
    \textsuperscript{4}Gaoling School of Artificial Intelligence, Renmin University of China\\
    \textsuperscript{5}Beijing Institute for General Artificial Intelligence \\
    \texttt{\{xl.zhao,zhaody\}@pku.edu.cn} \quad
    \texttt{\{redmondliu,shumingshi\}@tencent.com}\\
    \texttt{lucas.futingchen@gmail.com} \quad
    \texttt{ruiyan@ruc.edu.cn} \\
}

\begin{document}
\maketitle

\renewcommand{\thefootnote}{\fnsymbol{footnote}}
\footnotetext[1]{This work was done while X. Zhao was an intern at Tencent AI Lab.}
\footnotetext[2]{Corresponding authors: Dongyan Zhao and Rui Yan.}
\setcounter{footnote}{0}
\renewcommand{\thefootnote}{\arabic{footnote}}

\begin{abstract}
With the availability of massive general-domain dialogue data, pre-trained dialogue generation appears to be super appealing to transfer knowledge from the general domain to downstream applications. In most existing work, such transferable ability is mainly obtained by fitting a large model with hundreds of millions of parameters on massive data in an {\em exhaustive} way, leading to inefficient running and poor interpretability. This paper proposes a novel dialogue generation model with a latent structure that is easily transferable from the general domain to downstream tasks in a {\em lightweight} and {\em transparent} way. Experiments on two benchmarks validate the effectiveness of the proposed model. Thanks to the transferable latent structure, our model is able to yield better dialogue responses than four strong baselines in terms of both automatic and human evaluations, and our model with about 22\% parameters particularly delivers a 5x speedup in running time compared with the strongest baseline. Moreover, the proposed model is explainable by interpreting the discrete latent variables.

\end{abstract}

\section{Introduction}
Conversation between humans and machines has long been a goal of artificial intelligence (AI). Building an open-domain dialogue system with data-driven techniques has gotten a lot of attention in the AI and NLP fields in recent years, thanks to breakthroughs in deep learning~\cite{sutskever2014sequence,gehring2017convolutional,vaswani2017attention}. In particular, with the availability of massive human dialogue data (e.g., the Reddit comments) on social media~\cite{adiwardana2020towards}, pre-trained dialogue generation appears to be super appealing to alleviate potential discrepancies between general domain and downstream applications~\cite{zhang2020dialogpt,bao2019plato,bao2021plato2,li2021conversations}.

The common idea behind the pre-trained dialogue generation can be highlighted as a two-step pipeline: a) it firstly trains a deep neural model on massive general-domain dialogue data, b) and then transfers the model into downstream tasks via fine-tuning or zero-shot learning. Under this pipeline, the transferability is mainly obtained by fitting a large model with millions of parameters on massive data in an {\em exhaustive} way. Consequently, the downsides in existing works are obvious: their running is inefficient and their outputs are difficult to explain. 

This paper thereby aims to build a pre-trained dialogue model which is easily transferable from the general domain to downstream tasks in a {\em lightweight} and {\em transparent} way. To this end, we propose a novel dialogue model with a latent structure consisting of several latent variables. By using some self-supervised tasks to endow its latent variables with some prior properties during training, the latent structure makes the knowledge better transferable across different domains. Specifically, we first propose to incorporate the transformer architecture with a discrete conversation flow. Given a dialogue session, our model will sequentially infer the discrete state for each utterance which provides essential hints for future states and has an effect on the generation of the associated utterance. We further propose a method to disentangle the context-sensitive information from the conversation flow, which is achieved by two disentangled latent variables to capture the context-sensitive information (e.g., topic and persona) and the context-independent information (e.g., dialogue logic for each utterance) respectively. Through tailor-designed self-supervised tasks, the context-sensitive latent variable is able to capture the holistic information of a dialogue session while the context-independent variable is supposed to reflect the dynamic flow of dialogue in each utterance.
Meanwhile, the model is optimized with variational inference by maximizing the evidence lower bound of the likelihood.

We conduct experiments with two multi-turn dialogue generation benchmarks, including DailyDialog~\cite{li2017dailydialog} and ConvAI2~\cite{dinan2020second}. Thanks to the transferable latent structure, our model is able to yield better dialogue responses than four strong baselines in terms of both automatic and human evaluations, and our model including about 22\% - 66\% parameters particularly delivers a 2x - 30x speedup in running time. Moreover, the proposed model is explainable by visualizing the discrete latent variables. 

Our contributions in the paper are three-fold: (1) We present a context-free dialogue structure that captures the prior knowledge about state transition in a large-scale dialogue corpus. Furthermore, with the help of this dialogue structure, our model outperforms the state-of-the-art dialogue pre-training method with much fewer parameters. (2) We propose a disentangled structure learning framework to induce a context-free dialogue structure that enjoys better transferability and interpretability. (3) We empirically verify the effectiveness and efficiency of the proposed model on two benchmarks.

\section{Related Work}

The success of neural networks in machine translation promotes early research on end-to-end open-domain dialogue generation~\cite{ritter2011data,shangL2015neural,vinyals2015neural}. Various adaptations to the vanilla encoder-decoder architecture have been built to model the structure of dialogue contexts~\cite{serban2016building,serban2017hierarchical,zhang2019recosa}; improve response diversity~\cite{li2015diversity,zhao2017learning,tao2018get}; introduce external knowledge~\cite{dinan2018wizard,zhao2019low,zhao2020knowledge}; and control response qualities~\cite{xu2019neural,zhou2017emotional,zhang2018learning,wang2018learning,see2019makes}.

Large-scale pre-training for open-domain dialogue generation has recently become promising as a way to bridge the gap between conversation with existing systems and conversation with humans. Inspired by the successfulness of GPT-2~\cite{radford2019language}, \citet{zhang2020dialogpt} propose to train the transformer models on a very large dialogue dataset to generate informative text. \citet{bao2019plato} further use discrete latent variables to address the one-to-many mapping problem in open-domain dialogue. Despite prior successes, the dialogue context is simply concatenated as a long sequence, which may fail to capture the discourse-level coherence among utterances. To this end, \citet{gu2021dialogbert} and \citet{li2021conversations} introduce more self-supervision objectives to capture the discourse-level coherence and the dynamic information flow respectively.

The concept of dialogue structure has proven useful in modeling the complicated relationships between utterances. In the field of task-oriented dialogue, \citet{shi2019unsupervised} propose a discrete variational recurrent neural network (DVRNN) to learn the dialogue structure through unsupervised learning; \citet{qiu2020structured} further propose to enhance prior work with a structured attention mechanism; and \citet{sun2021unsupervised} propose a conversational graph to represent deterministic dialogue structure, where nodes and edges represent the utterance and context information, respectively. In the field of open-domain dialogue, \citet{xu2021discovering} construct a large dialogue structure graph with around $1.6$ million vertices to cover a wide range of topics. This work introduces a disentangled structure learning framework, which can induce a transferable substructure and an interpretable dialogue substructure, to incorporate the structural bias in dialogue pre-training. Thanks to the tailor-designed self-supervised tasks, our latent structure is more general than the dialogue structure in existing work.

\section{Approach}
\subsection{Overview}

Let $X=(u_1,u_2,\cdots, u_{n})$ denote a dialogue session, with $u_{t}=(w_{t,1},w_{t,2},\cdots,w_{t,m})$ denoting the $t$-th utterance and $w_{t,i}$ the $i$-th token in it. The number of utterances in a session and the number of tokens in each utterance are represented by $n$ and $m$, respectively. The conversational context for $u_t$ is $u_{<t}=(u_1, u_2, \cdots, u_{t-1})$. Our ultimate goal is to develop a generation model $p(u_t|u_{<t})$ that can predict the next utterance based on the context of the conversation.

\begin{figure}
\centering
\includegraphics[width=1.0\linewidth]{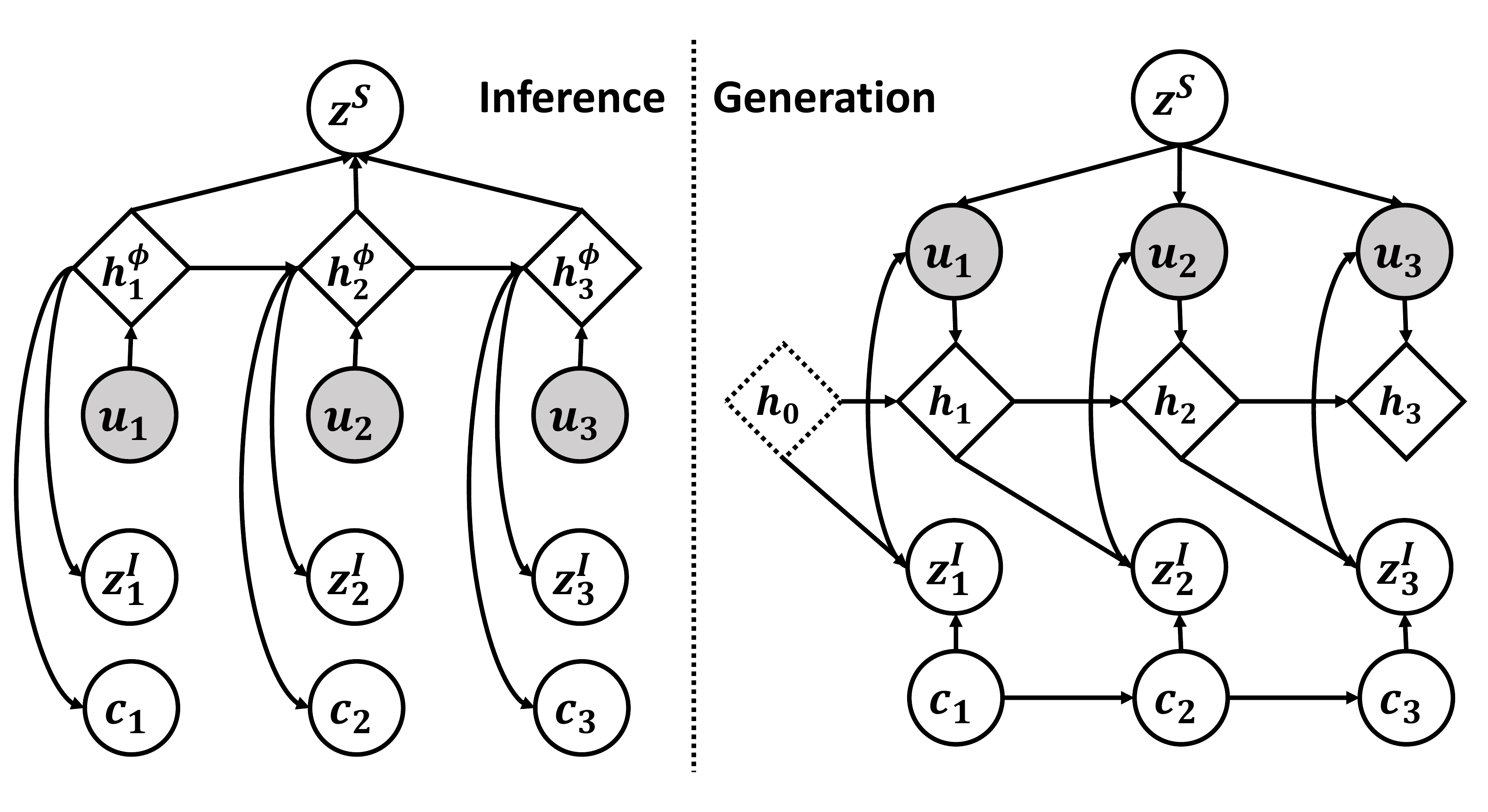}
\caption{Graphical illustrations of generation and inference processes. Left: inference of the approximate posterior as described in Section~\ref{sec:inference}. Right: generation of $u$ and computation of prior as described in Section~\ref{sec:generation}. The latent structure consists of $c$, $z^I$ and $z^S$. The initial hidden state $h_0$ is a trainable parameter.}
\label{fig:model}
\end{figure}

Figure~\ref{fig:model} illustrates the overview of our graphical model, which includes the proposed {\bf latent structure} consisting of three kinds of latent variables, i.e., $c=[c_1, c_2, \cdots, c_n]$, $z^I=[z^I_1, z^I_2, \cdots, z^I_n]$ and $z^{S}$. Specifically, $c$ depicts the flow of a conversation, and each $c_i\in \{1, \cdots, N\}$ is a discrete latent variable with $N$ as a hyper-parameter. It is worth noting that $c$ is designed for {\em interpretability}: by interpreting these discrete variables, humans are able to understand the logical flow of the conversation as to be shown in Section~\ref{sec:case}. Moreover, $z^{S}$ and $z^I$ are two disentangled latent variables to capture the context-sensitive information and context-independent information in a dialogue session respectively. In this way, through disentangling $z^{S}$ and $z^{I}$ with tailor-designed self-supervised learning objectives (as will be described in Section~\ref{sec:learn}), our model is able to capture intrinsic conversation flow for better generalization to different domains (i.e., {\em transferability}).

With our designed latent structure, given a conversational context $u_{<t}$, the generation of the next utterance $u_t$ can be roughly decomposed into two steps: (1) {\em infer} the conversation flow $[c_1, c_2, \cdots, c_{t-1}]$ and the context-sensitive variable $z^S$ based on context information, as shown in Figure~\ref{fig:model} (left). (2) compute the priors of $c_t$ and $z^I_t$, and then {\em generate} the next utterance $u_t$ with $z^I_t$ and $z^S$, as shown in Figure~\ref{fig:model} (right).

\subsection{Generation}
\label{sec:generation}
\paragraph{Context Encoding.}
We first obtain the contextualized representations of utterances through pre-trained language models (PLMs). Specifically, we exploit GPT-$2$~\cite{radford2019language}, which is pre-trained using the causal language modeling objective and achieves state-of-the-art results on a range of text generation tasks, as the backbone of our model. Note that our technical novelty lies in the proposal of a disentangled structure learning framework that injects a transferable dialogue structure into PLMs. Given a dialogue session $X=(u_1, u_2, \cdots, u_n)$, we first construct the input $I$ by concatenating all utterances as a single consecutive token sequence: 
\begin{equation}
I=\textrm{[BOS]} u_1 \textrm{[EOS]} u_2 \textrm{[EOS]} \ldots \textrm{[EOS]} u_n \textrm{[EOS]},
\end{equation}
where $\textrm{[BOS]}$ and $\textrm{[EOS]}$ are special tokens designed to separate sentences. The input $I$ is then fed into the PLM and the contextualized representation for $X$ is defined as the hidden states at the last layer: 
\begin{equation}
h_{1,1}, \cdots, h_{t,i}, \cdots, h_{n,m}=f_{trans}(I) \in \mathbb{R}^{mn \times d},
\end{equation}
where $f_{trans}(\cdot)$ denotes the transformer model~\cite{vaswani2017attention} and $h_{t,i} \in \mathbb{R}^d$ denotes the hidden state corresponding to token $w_{t,i}$. It's notable that we use uni-directional attention since the learning objectives are applied to all utterances (as will be illustrated in Section~\ref{sec:learn}) and a bi-directional architecture will leak the future information.

The vector representation of the $t$-th utterance is obtained through attentive pooling~\cite{wu2020attentive}, which is defined as follows:
\begin{equation}
\begin{aligned}
\label{eq:sent}
    h_{t} &= \sum_{j=1}^{m} \alpha_{t,i} h_{t,i}, \quad
    \alpha_{t,i} &= \frac{e^{q \cdot h_{t,i}}}{\sum_{i=1}^{m}e^{q \cdot h_{t,i}}},
\end{aligned}
\end{equation}
where $q \in \mathbb{R}^{d}$ is the attention query vector.

\paragraph{Prior of Discrete Latent Variable.}
The discrete latent variables $[c_1, c_2, \cdots, c_n]$ are used to automatically discover the structural representation in dialogues, which is beneficial to analyze how conversation flow from one utterance to the next one and promotes interpretability. We exclude the impact of $u_{<t}$ on $c_t$ since there is usually a domain discrepancy between the pre-trained and downstream data, which limits the transferability of the learned conversation flow. As a result, we directly model the influence of $c_{<t}$ on $c_t$ in the prior. We employ the transformer model with uni-directional attention to generate the contextualized representation of $c_{<t}$:
\begin{equation}
\small
\label{eq:discrete1}
h^{c}_{1}, \cdots, h^{c}_{t-1}=f_{c-trans}([c_1, \cdots, c_{t-1}]) \in \mathbb{R}^{(t-1) \times d}.
\end{equation}
Then the probability of predicting $c_t$ is defined as:
\begin{equation}
\label{eq:discrete2}
    p(c_t|c_{<t}) = \textrm{Softmax}(f_{c-mlp}(h^{c}_{t-1})),
\end{equation}
where $f_{c-mlp}(\cdot)$ denotes a MLP network. Different from~\citet{shi2019unsupervised}, our model preserves the capacity to represent $n$-gram transition probability, which is superior for capturing long-term dependency in the conversation flow of open-domain dialogues.

\paragraph{Priors of Context-Sensitive and Context-Independent Variables.}
Despite the fact that the discrete latent variables can intuitively characterize conversation flow, the complexity of open-domain conversation necessitates a large number of dialogue states to address fine-grained semantics~\cite{xu2021discovering}, making model training highly challenging and resulting in poor generalization capacity. To alleviate the aforementioned difficulties, we introduce two latent variables to decouple the conversation flow and contextual information, namely the context-sensitive latent variable $z^S$ and the context-independent latent variable $z^I$.

The prior of context-sensitive latent variable $z^{S}$ is defined as a standard Gaussian distribution:
\begin{equation}
    p(z^{S})=\mathcal{N}(0, \mathbf{I}),
\end{equation}
where $\mathbf{I}$ denotes the unit matrix.

The context-independent latent variable is responsible for capturing dynamic information in each utterance. To achieve this, we condition the prior of $z^I_t$ on both contextualized representation of the previous utterance and the predicted discrete state for the current utterance:
\begin{equation}
\begin{aligned}
    p(z^{I}_{t}|u_{<t},c_{t}) &= \mathcal{N}({\mu}^{I}_{t},{\sigma}^{I}_{t} \mathbf{I}), \\
    {\mu}^{I}_{t}, {\sigma}^{I}_{t} &= f_{I-mlp}([h_{t-1};e(c_t)]),
\end{aligned}
\end{equation}
where $f_{I-mlp}(\cdot)$ denotes a MLP network, $e(\cdot)$ is the embedding of a latent state and $[\cdot;\cdot]$ denotes vector concatenation. We employ the Gumbel trick~\cite{jang2017categorical} to handle the discrete and undifferentiable process of sampling $c_t$.

\paragraph{Decoding.}
Given the contextualized representation $h_{t,i}$ for token $w_{t,i}$, the original GPT-$2$ model calculates the pre-softmax logit vector through a linear head, i.e., $p_{t,i}=\mathbf{W}_v h_{t,i}$, where $\mathbf{W}_v$ is a learnable parameter. To explicitly guide the generation through the context-independent latent variable $z^I_{t}$ and the context-sensitive latent variable $z^S$, we first project them into the space of $h_{t,i}$ and then calculate two pre-softmax logit vectors similar to $p_{t,i}$:
\begin{equation}
\begin{aligned}
    p^{I}_{t} = \mathbf{W}_v \mathbf{W}^{I}_v z^I_{t}, \quad
    p^{S} = \mathbf{W}_v \mathbf{W}^{S}_v z^{S},
\end{aligned}
\end{equation}
where $\mathbf{W}^{I}_v$ and $\mathbf{W}^{S}_v$ are learnable parameters. We employ the reparameterization trick~\cite{kingma2013auto} to allow gradient passing through the sampling of $z^{I}_t$ and $z^S$. The probability of generating the next token is then defined as:
\begin{equation}
\small
\begin{aligned}
    p(w_{t,i+1}|u_{<t}, w_{t,<i+1}, z^{I}_t, z^{S}) = \textrm{Softmax}(p_{t,i} + p^I_t + p^S).
\end{aligned}
\end{equation}

The parameterization of the generative model results in the following factorization:
\begin{equation}
\small
\begin{aligned}
    &p(u_{\leq n}, c_{\leq n}, z^{I}_{\leq n}, z^S) \\
    &= p(z^S)\prod_{t=1}^{n} \left(p(u_t|u_{<t}, z^I_t, z^S) p(z^I_t|u_{<t},c_t) p(c_t|c_{<t}) \right), 
\end{aligned}
\end{equation}
where the probability of generating $u_t$ is formulated as: $p(u_t|u_{<t},z^{I}_t, z^{S})=\prod_{i=1}^{m}p(w_{t,i}|u_{<t}, w_{t,<i}, z^{I}_t, z^{S})$.

\subsection{Inference}
\label{sec:inference}
For the inference of latent variables, we employ a lightweight transformer that is initialized by the first $6$ layers of the GPT-$2$ model. The vector representation of utterance $u_t$ is denoted as $h^{\phi}_t$ and is defined in the same way as Eq.\ref{eq:sent}. We introduce three auxiliary distributions $q_{\phi}(c|X)$, $q_{\phi}(z^{I}|X)$ and $q_{\phi}(z^{S}|X)$ which approximate to the posteriors of the discrete latent variable $c$, the context-independent latent variable $z^{I}$ and the context-sensitive latent variable $z^{S}$ respectively. 

Since the context-sensitive latent variable $z^S$ captures the holistic information about the whole session, we construct the posterior distribution $q_{\phi}(z^{S}|X)$ by summarizing the representations for each utterance:
\begin{equation}
\begin{aligned}
    q_{\phi}(z^{S}|X) &= \mathcal{N}(\hat{\mu}^{S},\hat{\sigma}^{S} \mathbf{I}), \\
    \hat{\mu}^{S}, \hat{\sigma}^{S} &= f_{S-mlp}^{\prime}(h^{\phi}),
\end{aligned}
\end{equation}
where $h^{\phi}=\frac{1}{n}\sum_{t=1}^{n}h^{\phi}_{t}$ and $f_{S-mlp}^{\prime}(\cdot)$ denotes a MLP network. Similarly, the posterior distribution $q_{\phi}(z^{I}|X)$ is defined as:
\begin{equation}
\begin{aligned}
    q_{\phi}(z^I|X) &= \prod_{t=1}^{n}q_{\phi}(z^I_t|X), \\
    q_{\phi}(z^{I}_{t}|X) &= \mathcal{N}(\hat{\mu}^{I}_{t},\hat{\sigma}^{I}_{t} \mathbf{I}), \\
    \hat{\mu}^{I}_{t}, \hat{\sigma}^{I}_{t} &= f_{I-mlp}^{\prime}(h^{\phi}_{t}),
\end{aligned}
\end{equation}
where $h^{\phi}_{t}$ encodes the contextual information of $u_t$. The posterior distribution $q_{\phi}(c|X)$ could be factorized as $\prod_{t=1}^{n} q_{\phi}(c_{t}|X)$ where $q_{\phi}(c_{t}|X)$ is a Categorical distribution parameterized by $\textrm{Softmax}(f_{c-mlp}^{\prime}(h^{\phi}_t))$. The inference process is depicted in Figure~\ref{fig:model}.

\subsection{Learning}
\label{sec:learn}

The log-likelihood of the conversation session $X$ is maximized using variational approximation, yielding the evidence lower bound objective (ELBO)~\cite{hoffman2013stochastic}:
\begin{equation}
\small
\begin{aligned}
    \mathcal{L}_{ELBO} =
    &\sum_{t=1}^{n}\mathbb{E}_{q_{\phi}(z^S|X) q_{\phi}(z^I_{t}|X)} \log p(u_t|u_{<t},z^I_t, z^S) \\
    & -\sum_{t=1}^{n} D_{\mathrm{KL}}\left(q_{\phi}(c_t|X) \| p(c_t|c_{<t})\right) \\
    & -\sum_{t=1}^{n} \mathbb{E}_{q_{\phi}(c_t)|X} D_{\mathrm{KL}}\left(q_{\phi}(z^{I}_{t} | X) \| p(z^{I}_{t}|u_{<t},c_t)\right) \\
    & -D_{\mathrm{KL}}\left(q_{\phi}(z^S|X) \| p(z^S)\right),
\end{aligned}
\end{equation}
where $D_{\mathrm{KL}}(\cdot \| \cdot)$ refers to Kullback–Leibler divergence. Detailed derivations are presented in Appendix~\ref{sec:app-elbo}.

In addition to optimizing the ELBO objective, we also exploit disentanglement to distill the holistic information into the context-sensitive latent variable while keeping the dynamic information in the context-independent latent variable and the discrete latent variable as demonstrated in the rest of this section.

\paragraph{Holistic Information Discrimination.}
The latent variable $z^{S}$ is supposed to only focus on the holistic information that refers to the time-invariant factors (e.g., interlocutor persona) and remains consistent throughout the whole dialogue session. To achieve this, we design a self-supervised task to eliminate the dynamic information from $z^{S}$. Ideally, the holistic information of a session $X$ is insensitive to a random shuffle of its internal utterances (i.e., $X_{shuf}$), but varies across randomly picked different dialogue sessions (i.e., $X_{neg}$). Thus we could maximize the following objective:

\begin{equation}
    \mathcal{L}_{HID} = \log \frac{e^{f_{sim}(z^{S}, z^{S}_{shuf}))}}{e^{f_{sim}(z^{S}, z^{S}_{shuf})} + e^{f_{sim}(z^{S}, z^{S}_{neg})}},
\end{equation}
where $z^{S}_{shuf}$ and $z^{S}_{neg}$ denote the context-sensitive latent variables of $X_{shuf}$ and $X_{neg}$ respectively, and $f_{sim}(\cdot, \cdot)$ is implemented as the cosine similarity between two vectors.

\paragraph{Dynamic Information Restoration.}
Since each utterance contains some utterance-specific features that are independent of the conversational context, it is reasonable to encourage the context-independent latent variable $z^{I}$ to be aware of the dynamic information flow in a session. Specifically, we design a surrogate task to recover the verbs in an utterance $u_t$ given the corresponding context-independent latent variable $z^{I}_{t}$:
\begin{equation}
    \mathcal{L}_{DIR} = \sum_{t=1}^{n}\sum_{i=1}^{m} \delta_{t,i} \log p(w_{t,i} | z^{I}_{t}),
\end{equation}
where $\delta_{t,i}=1$ if the token $w_{t,i}$ is a verb, otherwise $\delta_{t,i}=0$.  $p(w_{t,i}|z^{I}_{t})=\textrm{Softmax}(\mathbf{W}_{verb} z^{I}_{t})$ outputs a probability distribution over all verbs in the vocabulary with $\mathbf{W}_{verb}$ a learnable parameter.

\paragraph{Mutual Information Minimization.}
Since the task of Dynamic Information Restoration can not guarantee that the holistic information is exclusive in $z^{I}$, we further introduce the mutual information objective as a regularization to minimize the relationship between $z^{S}$ and $z^{I}$:
\begin{equation}
    \mathcal{L}_{MIM} = -\sum_{t=1}^{n}[H(z^{S})+H(z_{t}^{I})-H(z^{S}, z_{t}^{I})],
\end{equation}
where $H(\cdot)$ denotes the entropy which is estimated through minibatch-weighted sampling~\cite{chen2019isolating,zhu2020s3vae}:

\begin{equation}
\small
\begin{aligned}
    H(z) &\equiv -\mathbb{E}_{q(z)}[\log q(z)]  \\
    &\approx -\frac{1}{B} \sum_{i=1}^{B}\left[\log \frac{1}{M B} \sum_{j=1}^{B} q(z(u^{(i)}_{t}) | u^{(j)}_{t})\right],
\end{aligned}
\end{equation}
for $z=z^{I}_{t}$, $z^{S}$ or $(z^{S}, z^{I}_{t})$, where $u^{(i)}_{t}$ denotes $u_t$ in the $i$-th data point,  $z(u^{(i)}_{t})$ is a sample from $q(z|u^{(i)}_t)$, $M$ and $B$ are the data size and minibatch size respectively. 

The final learning objective is defined as:
\begin{equation}
    \mathcal{L} = \mathcal{L}_{ELBO} + \alpha (\mathcal{L}_{HID} + \mathcal{L}_{DIR} + \mathcal{L}_{MIM}),
\end{equation}
where $\alpha$ is a hyper-parameter to balance the objective of evidence lower bound and those related to disentanglement.

\begin{table*}[t]
\centering
\resizebox{0.85\linewidth}{!}{
\begin{tabular}{lcccccccccc}
\toprule
\multicolumn{1}{c}{Models}     & \multicolumn{1}{c}{BLEU-1}         & \multicolumn{1}{c}{BLEU-2}         & \multicolumn{1}{c}{BLEU-3}         & \multicolumn{1}{c}{BLEU-4}         & \multicolumn{1}{c}{ROUGE-1}        & \multicolumn{1}{c}{ROUGE-2}        & \multicolumn{1}{c}{ROUGE-L}        & \multicolumn{1}{c}{METEOR}         & \multicolumn{1}{c}{Distinct-1}    & Distinct-2     \\ \midrule
\multicolumn{11}{c}{Zero-resource setting}                                                                                                                                                                                                                                                                                                                                                                                         \\ \midrule
\multicolumn{1}{l}{DialoGPT}   & \multicolumn{1}{c}{4.29}           & \multicolumn{1}{c}{1.52}           & \multicolumn{1}{c}{0.74}           & \multicolumn{1}{c}{0.41}           & \multicolumn{1}{c}{9.01}           & \multicolumn{1}{c}{2.12}           & \multicolumn{1}{c}{8.63}           & \multicolumn{1}{c}{3.93}           & \multicolumn{1}{c}{6.97}          & 31.79          \\ 
\multicolumn{1}{l}{DialogBERT} &  \multicolumn{1}{c}{6.91}           & \multicolumn{1}{c}{1.22}           & \multicolumn{1}{c}{0.22}           & \multicolumn{1}{c}{0.02}           & \multicolumn{1}{c}{5.54}           & \multicolumn{1}{c}{0.21}           & \multicolumn{1}{c}{4.87}           & \multicolumn{1}{c}{4.35}           & \multicolumn{1}{c}{5.12}          & 30.84          \\ 
\multicolumn{1}{l}{PLATO-2}    & \multicolumn{1}{c}{5.38}           & \multicolumn{1}{c}{1.86}           & \multicolumn{1}{c}{0.85}           & \multicolumn{1}{c}{0.47}           & \multicolumn{1}{c}{8.90}           & \multicolumn{1}{c}{2.08}           & \multicolumn{1}{c}{8.49}           & \multicolumn{1}{c}{3.72}           & \multicolumn{1}{c}{4.88}          & 19.50          \\ 
\multicolumn{1}{l}{DialoFlow}  & \multicolumn{1}{c}{4.76}           & \multicolumn{1}{c}{1.40}           & \multicolumn{1}{c}{0.52}           & \multicolumn{1}{c}{0.20}           & \multicolumn{1}{c}{9.07}           & \multicolumn{1}{c}{1.36}           & \multicolumn{1}{c}{8.52}           & \multicolumn{1}{c}{5.08}           & \multicolumn{1}{c}{5.35}          & 20.23          \\ \midrule
\multicolumn{1}{l}{Ours}       & \multicolumn{1}{c}{\textbf{9.58}}  & \multicolumn{1}{c}{\textbf{3.40}}  & \multicolumn{1}{c}{\textbf{1.64}}  & \multicolumn{1}{c}{\textbf{0.90}}  & \multicolumn{1}{c}{\textbf{11.99}} & \multicolumn{1}{c}{\textbf{2.50}}  & \multicolumn{1}{c}{\textbf{11.35}} & \multicolumn{1}{c}{\textbf{5.31}}  & \multicolumn{1}{c}{5.11}          & 23.86          \\ \bottomrule
\multicolumn{11}{c}{Full-resource setting}                                                                                                                                                                                                                                                                                                                                                                                         \\ \midrule
\multicolumn{1}{l}{DialoGPT}   &  \multicolumn{1}{c}{18.97}          & \multicolumn{1}{c}{8.67}           & \multicolumn{1}{c}{4.95}           & \multicolumn{1}{c}{3.09}           & \multicolumn{1}{c}{15.97}          & \multicolumn{1}{c}{4.26}           & \multicolumn{1}{c}{14.38}          & \multicolumn{1}{c}{9.81}           & \multicolumn{1}{c}{2.71}          & 15.75          \\ 
\multicolumn{1}{l}{DialogBERT} &  \multicolumn{1}{c}{11.27}          & \multicolumn{1}{c}{2.20}           & \multicolumn{1}{c}{0.62}           & \multicolumn{1}{c}{0.18}           & \multicolumn{1}{c}{4.56}           & \multicolumn{1}{c}{0.23}           & \multicolumn{1}{c}{4.21}           & \multicolumn{1}{c}{4.20}           & \multicolumn{1}{c}{3.91}          & 24.86          \\ 
\multicolumn{1}{l}{PLATO-2}    &  \multicolumn{1}{c}{14.10}          & \multicolumn{1}{c}{6.08}           & \multicolumn{1}{c}{3.32}           & \multicolumn{1}{c}{2.04}           & \multicolumn{1}{c}{14.48}          & \multicolumn{1}{c}{3.74}           & \multicolumn{1}{c}{13.49}          & \multicolumn{1}{c}{6.34}           & \multicolumn{1}{c}{2.64}          & 10.31          \\ 
\multicolumn{1}{l}{DialoFlow}  &  \multicolumn{1}{c}{17.78}          & \multicolumn{1}{c}{8.51}           & \multicolumn{1}{c}{5.21}           & \multicolumn{1}{c}{3.56}           & \multicolumn{1}{c}{18.14}          & \multicolumn{1}{c}{6.06}           & \multicolumn{1}{c}{16.54}          & \multicolumn{1}{c}{10.14}          & \multicolumn{1}{c}{2.82}          & 18.15          \\ \midrule
\multicolumn{1}{l}{Ours}       &  \multicolumn{1}{c}{\textbf{24.87}} & \multicolumn{1}{c}{\textbf{16.70}} & \multicolumn{1}{c}{\textbf{13.69}} & \multicolumn{1}{c}{\textbf{12.16}} & \multicolumn{1}{c}{\textbf{22.29}} & \multicolumn{1}{c}{\textbf{11.89}} & \multicolumn{1}{c}{\textbf{21.47}} & \multicolumn{1}{c}{\textbf{11.36}} & \multicolumn{1}{c}{\textbf{4.88}} & \textbf{28.28} \\ \bottomrule
\end{tabular}
}
\vspace{-1mm}
\caption{Automatic evaluation results on the test set of DailyDialog. Numbers in bold mean that the improvement to the best-performing baseline is statistically significant (t-test with p-value$<0.05$).}
\vspace{-1mm}
\label{tab:exp-daily}
\end{table*}

\begin{table*}[t]
\centering
\resizebox{0.85\linewidth}{!}{
\begin{tabular}{lcccccccccc}
\toprule
\multicolumn{1}{c}{Models}     & \multicolumn{1}{c}{BLEU-1}         & \multicolumn{1}{c}{BLEU-2}        & \multicolumn{1}{c}{BLEU-3}        & \multicolumn{1}{c}{BLEU-4}        & \multicolumn{1}{c}{ROUGE-1}        & \multicolumn{1}{c}{ROUGE-2}       & \multicolumn{1}{c}{ROUGE-L}        & \multicolumn{1}{c}{METEOR}         & \multicolumn{1}{c}{Distinct-1}    & Distinct-2    \\ \midrule
\multicolumn{11}{c}{Zero-resource setting}                                                                                                                                                                                                                                                                                                                                                                                    \\ \midrule
\multicolumn{1}{l}{DialoGPT}   & \multicolumn{1}{c}{7.23}           & \multicolumn{1}{c}{3.14}          & \multicolumn{1}{c}{1.64}          & \multicolumn{1}{c}{0.90}          & \multicolumn{1}{c}{12.90}          & \multicolumn{1}{c}{2.79}          & \multicolumn{1}{c}{12.40}          & \multicolumn{1}{c}{4.99}           & \multicolumn{1}{c}{6.26}          & 31.60         \\ 
\multicolumn{1}{l}{DialogBERT} & \multicolumn{1}{c}{6.53}           & \multicolumn{1}{c}{1.24}          & \multicolumn{1}{c}{0.27}          & \multicolumn{1}{c}{0.09}          & \multicolumn{1}{c}{6.15}           & \multicolumn{1}{c}{0.22}          & \multicolumn{1}{c}{5.48}           & \multicolumn{1}{c}{4.54}           & \multicolumn{1}{c}{5.10}          & 30.75         \\ 
\multicolumn{1}{l}{PLATO-2}    & \multicolumn{1}{c}{7.85}           & \multicolumn{1}{c}{3.38}          & \multicolumn{1}{c}{1.78}          & \multicolumn{1}{c}{1.02}          & \multicolumn{1}{c}{11.04}          & \multicolumn{1}{c}{2.71}          & \multicolumn{1}{c}{10.58}          & \multicolumn{1}{c}{4.73}           & \multicolumn{1}{c}{4.49}          & 18.94         \\ 
\multicolumn{1}{l}{DialoFlow}  & \multicolumn{1}{c}{8.17}           & \multicolumn{1}{c}{3.27}          & \multicolumn{1}{c}{1.43}          & \multicolumn{1}{c}{0.63}          & \multicolumn{1}{c}{12.01}          & \multicolumn{1}{c}{2.54}          & \multicolumn{1}{c}{11.22}          & \multicolumn{1}{c}{5.74}           & \multicolumn{1}{c}{6.92}          & 27.95         \\ \midrule
\multicolumn{1}{l}{Ours}       & \multicolumn{1}{c}{\textbf{11.16}} & \multicolumn{1}{c}{\textbf{4.62}} & \multicolumn{1}{c}{\textbf{2.34}} & \multicolumn{1}{c}{\textbf{1.29}} & \multicolumn{1}{c}{\textbf{14.73}} & \multicolumn{1}{c}{\textbf{3.25}} & \multicolumn{1}{c}{\textbf{14.11}} & \multicolumn{1}{c}{\textbf{5.82}}  & \multicolumn{1}{c}{5.24}          & 27.79         \\ \bottomrule
\multicolumn{11}{c}{Full-resource setting}                                                                                                                                                                                                                                                                                                                                                                                    \\ \midrule
\multicolumn{1}{l}{DialoGPT}   &  \multicolumn{1}{c}{14.92}          & \multicolumn{1}{c}{7.32}          & \multicolumn{1}{c}{4.01}          & \multicolumn{1}{c}{2.34}          & \multicolumn{1}{c}{17.07}          & \multicolumn{1}{c}{4.61}          & \multicolumn{1}{c}{15.53}          & \multicolumn{1}{c}{10.78}          & \multicolumn{1}{c}{1.28}          & 7.33          \\ 
\multicolumn{1}{l}{DialogBERT} &  \multicolumn{1}{c}{12.69}          & \multicolumn{1}{c}{3.11}          & \multicolumn{1}{c}{0.92}          & \multicolumn{1}{c}{0.34}          & \multicolumn{1}{c}{8.20}           & \multicolumn{1}{c}{0.54}          & \multicolumn{1}{c}{7.55}           & \multicolumn{1}{c}{4.66}           & \multicolumn{1}{c}{1.03}          & 5.67          \\ 
\multicolumn{1}{l}{PLATO-2}    &  \multicolumn{1}{c}{15.21}          & \multicolumn{1}{c}{7.25}          & \multicolumn{1}{c}{4.11}          & \multicolumn{1}{c}{2.63}          & \multicolumn{1}{c}{13.90}          & \multicolumn{1}{c}{4.07}          & \multicolumn{1}{c}{13.13}          & \multicolumn{1}{c}{6.94}           & \multicolumn{1}{c}{1.56}          & 5.94          \\ 
\multicolumn{1}{l}{DialoFlow}  &  \multicolumn{1}{c}{18.57}          & \multicolumn{1}{c}{8.83}          & \multicolumn{1}{c}{4.69}          & \multicolumn{1}{c}{2.70}          & \multicolumn{1}{c}{18.35}          & \multicolumn{1}{c}{4.64}          & \multicolumn{1}{c}{16.95}          & \multicolumn{1}{c}{9.76}           & \multicolumn{1}{c}{1.85}          & 9.96          \\ \midrule
\multicolumn{1}{l}{Ours}       &  \multicolumn{1}{c}{\textbf{20.15}} & \multicolumn{1}{c}{\textbf{9.41}} & \multicolumn{1}{c}{\textbf{5.17}} & \multicolumn{1}{c}{\textbf{3.09}} & \multicolumn{1}{c}{\textbf{19.19}} & \multicolumn{1}{c}{\textbf{5.04}} & \multicolumn{1}{c}{\textbf{17.38}} & \multicolumn{1}{c}{10.13} & \multicolumn{1}{c}{1.69} & 8.74 \\ \bottomrule
\end{tabular}
}
\vspace{-1mm}
\caption{Automatic evaluation results on the test set of ConvAI2. Numbers in bold mean that the improvement to the best-performing baseline is statistically significant (t-test with p-value $<0.05$).}
\vspace{-1mm}
\label{tab:exp-convai}
\end{table*}

\section{Experimental Setup}
\subsection{Datasets}
We follow~\citet{zhang2020dialogpt} to adopt the Reddit comments as our pre-training data. We evaluate our model on two benchmark datasets for multi-turn dialogue generation, including DailyDialog~\cite{li2017dailydialog} and ConvAI2~\cite{dinan2020second}. More details about all datasets are provided in Appendix~\ref{sec:app-data}.

\subsection{Evaluation Metrics}
\paragraph{Automatic Evaluation.}

We choose three commonly used reference-based metrics including BLEU~\cite{papineni2002bleu}, METEOR~\cite{lavie2007meteor} and ROUGE~\cite{lin2004rouge}, where BLEU and METEOR are computed with an open source NLG evaluation tool available at \url{https://github.com/Maluuba/nlg-eval}, and ROUGE is calculated with the code published at \url{https://github.com/bckim92/language-evaluation}. We report the F1 scores for ROUGE-$1$, ROUGE-$2$ and ROUGE-L. We also use Distinct~\cite{li2015diversity} to evaluate the lexical diversity with Distinct-1 and Distinct-2 denoting ratios of distinct unigrams and bigrams in responses, respectively.

\begin{table*}[t]
\centering
\resizebox{0.85\linewidth}{!}{
\begin{tabular}{lcccccccccc}
\toprule
\multicolumn{1}{c}{\multirow{2}{*}{Models}} & \multicolumn{5}{c}{DailyDialog}                                                                                                               & \multicolumn{5}{c}{ConvAI2}                                                                                                                    \\ 
\cmidrule(lr){2-6}\cmidrule(lr){7-11}
\multicolumn{1}{c}{}                        & \multicolumn{1}{c}{Fluency} & \multicolumn{1}{c}{Relevance} & \multicolumn{1}{c}{Informativeness} & \multicolumn{1}{c}{Engagement} & Kappa & \multicolumn{1}{c}{Fluency} & \multicolumn{1}{c}{Relevance} & \multicolumn{1}{c}{Informativeness} & \multicolumn{1}{c}{Engagement} & Kappa \\ \midrule
DialoGPT                                     & \multicolumn{1}{c}{1.67}    & \multicolumn{1}{c}{1.71}      & \multicolumn{1}{c}{1.69}            & \multicolumn{1}{c}{1.75}       & 0.61  & \multicolumn{1}{c}{1.73}    & \multicolumn{1}{c}{1.65}      & \multicolumn{1}{c}{1.63}            & \multicolumn{1}{c}{1.79}       & 0.74  \\ 
DialogBERT                                   & \multicolumn{1}{c}{1.43}    & \multicolumn{1}{c}{1.39}      & \multicolumn{1}{c}{1.33}            & \multicolumn{1}{c}{1.37}       & 0.77  & \multicolumn{1}{c}{1.36}    & \multicolumn{1}{c}{1.35}      & \multicolumn{1}{c}{1.31}            & \multicolumn{1}{c}{1.27}       & 0.75  \\ 
PLATO-2                                      & \multicolumn{1}{c}{1.74}    & \multicolumn{1}{c}{1.68}      & \multicolumn{1}{c}{1.73}            & \multicolumn{1}{c}{1.65}       & 0.70  & \multicolumn{1}{c}{1.72}    & \multicolumn{1}{c}{1.64}      & \multicolumn{1}{c}{1.72}            & \multicolumn{1}{c}{1.70}       & 0.64  \\ 
DialoFlow                                    & \multicolumn{1}{c}{1.71}    & \multicolumn{1}{c}{1.72}      & \multicolumn{1}{c}{1.71}            & \multicolumn{1}{c}{1.76}       & 0.65  & \multicolumn{1}{c}{1.79}    & \multicolumn{1}{c}{1.70}      & \multicolumn{1}{c}{1.78}            & \multicolumn{1}{c}{1.81}       & 0.69  \\ \midrule
Ours                                         & \multicolumn{1}{c}{\textbf{1.81}}    & \multicolumn{1}{c}{\textbf{1.79}}      & \multicolumn{1}{c}{1.74}            & \multicolumn{1}{c}{\textbf{1.86}}       & 0.68  & \multicolumn{1}{c}{1.83}    & \multicolumn{1}{c}{\textbf{1.76}}      & \multicolumn{1}{c}{1.80}            & \multicolumn{1}{c}{\textbf{1.89}}       & 0.70  \\ \bottomrule
\end{tabular}
}
\vspace{-1mm}
\caption{Human evaluation results on DailyDialog and ConvAI2. Numbers in bold mean that the improvement to the best-performing
baseline is statistically significant (t-test with p-value $<0.05$). }
\vspace{-1mm}
\label{tab:exp-human}
\end{table*}

\paragraph{Human Evaluation.}

We randomly sample $300$ examples from the test sets of DailyDialog and ConvAI2 respectively, and recruit $6$ well-educated native speakers to do qualitative analysis on the responses generated by our model and all competitive baselines, which are randomly shuffled to hide identification. The annotators judge the quality of the responses from four aspects: 
(1) \emph{Fluency}: whether the response is fluent without any grammatical errors; 
(2) \emph{Relevance}: whether the response is coherent with the context; 
(3) \emph{Informativeness}: whether the response contains informative content; 
(4) \emph{Engagement}: how much does the annotator like the response. Each annotator assigns a score from $\{0, 1, 2\}$ (representing ``bad'', ``fair'' and ``good'' respectively) to each response for each aspect. Each response obtains four scores for the aforementioned four aspects, and the agreement among all annotators is measured via Fleiss' kappa~\cite{fleiss1971measuring}.

\subsection{Baseline Models}
The following models are selected as baselines: 
(1) \textbf{DialoGPT.} A model that is pre-trained on the Reddit comments and attains a performance close to human in single-turn dialogues~\cite{zhang2020dialogpt}. We adopt the \textit{medium}-sized model which achieves the best performance in the original paper.  
(2) \textbf{DialogBERT.} A model that encodes the dialogue context with a hierarchical transformer architecture~\cite{gu2021dialogbert}.  
(3) \textbf{PLATO-2.} A model that learns a fine-grained one-to-many generation with the advent of a discrete latent variable~\cite{bao2021plato2}. It is notable that the performance of PLATO-2 is superior to PLATO~\cite{bao2019plato} by introducing more parameters and training data.
(4) \textbf{DialoFlow.} A model that is pre-trained on the Reddit comments and incorporates a dynamic flow mechanism to model the context flow in dialogues~\cite{li2021conversations}. We adopt the \textit{large}-sized model which achieves the best performance in the original paper. 

All the baselines are taken from their open-source implementations. We continue to train DialogBERT and PLATO-2 on the Reddit comments for the sake of fairness. The parameter sizes of all baselines are shown in Table \ref{tab:exp-speed}. We provide more implementation details in Appendix~\ref{sec:app-imp}.

\section{Results and Discussion}
\subsection{Main Results}
In this section, we will compare the performance of various models on DailyDialog and ConvAI2, as well as provide some further analyses. We conduct experiments in two different settings, including zero-resource and full-resource, both of which are commonly employed by pre-trained language models. All models solely use the Reddit comments during training in the zero-resource scenario, however, in the full-resource situation, all models are pre-trained on the Reddit comments and then fine-tuned on downstream tasks. Table \ref{tab:exp-daily} and Table \ref{tab:exp-convai} show the performance of our model on DailyDialog and ConvAI2 respectively.
From the results, we can observe that:
(1) Although our model is much smaller than the other baseline models, it achieves the best performance on appropriateness-related metrics (i.e., BLEU, ROUGE and METEOR) and performs comparably on distinctness-related metrics (i.e., Distinct) at the same time, demonstrating the effectiveness of a context-free dialogue structure. Additionally, our model takes advantage of $z^S$ and $z^I$ to capture both the time-invariant and time-varying factors and generate a coherent response.
(2) DialoFlow outperforms DialoGPT on most metrics after fine-tuning, but not as good as ours. This verifies the necessity of capturing the dialogue flow in PLMs, and the proposed context-free dialogue structure is more competent.
(3) On the DailyDialog, our model outperforms baselines by a larger margin than that on the ConvAI2.
This is possibly due to the introduction of dialogue act flow in the construction of DailyDialog, which has a similar effect to the dialogue structure.

\paragraph{Human Evaluation.}
Table \ref{tab:exp-human} shows the results of the human evaluation. While our model achieves comparable performance to the others in terms of \emph{Fluency} and \emph{Informativeness}, it outperforms them on both \emph{Relevance} and \emph{Engagement}, agreeing with the results of automatic evaluation. All kappa values are more than $0.6$, indicating that the annotators are in agreement.

\begin{table}[h!]
\centering
\resizebox{1.0\linewidth}{!}{
\begin{tabular}{lccccc}
\toprule
\multicolumn{1}{c}{\multirow{3}{*}{Models}} & \multirow{3}{*}{Parameter Size} & \multicolumn{4}{c}{Decoding Speed (ms)}                                                              \\ \cmidrule(lr){3-6}
\multicolumn{1}{c}{}                        &                                 & \multicolumn{2}{c}{DailyDialog}                            & \multicolumn{2}{c}{ConvAI2}           \\ \cmidrule(lr){3-4}\cmidrule(lr){5-6}
\multicolumn{1}{c}{}                        &                                 & \multicolumn{1}{c}{zero}    & \multicolumn{1}{c}{full}    & \multicolumn{1}{c}{zero}    & full    \\ \midrule
DialoGPT                                     & 345M                            & \multicolumn{1}{c}{34.93}  & \multicolumn{1}{c}{36.72}  & \multicolumn{1}{c}{39.01}  & 39.45  \\ 
DialogBERT                                   & 338M                            & \multicolumn{1}{c}{56.10}  & \multicolumn{1}{c}{104.96} & \multicolumn{1}{c}{104.66} & 49.47  \\ 
PLATO-2                                      & 310M                            & \multicolumn{1}{c}{482.02} & \multicolumn{1}{c}{401.19} & \multicolumn{1}{c}{479.39} & 347.88 \\ 
DialoFlow                                    & 941M                            & \multicolumn{1}{c}{94.80} & \multicolumn{1}{c}{96.10} & \multicolumn{1}{c}{103.29} & 82.94 \\ \midrule
Ours                                         & 207M                            & \multicolumn{1}{c}{15.45}  & \multicolumn{1}{c}{17.34}  & \multicolumn{1}{c}{17.13}  & 16.72  \\ \bottomrule
\end{tabular}
}
\vspace{-1mm}
\caption{Evaluation results about decoding speed. ``zero'' and ``full'' are abbreviations for zero-resource setting and full-resource setting respectively.}
\vspace{-1mm}
\label{tab:exp-speed}
\end{table}

\paragraph{Speed Test.}
We further compare our model with baselines in terms of decoding speed. Specifically, we calculate the average prediction time per word in response generation in both zero-resource and full-resource settings utilizing all dialogues in the test sets. The experiments are conducted on an RTX 3090. Table \ref{tab:exp-speed} shows the speed comparison results. The discrete variable $c$ learns a general transition pattern from the entire corpus, which compensates for the small parameter scale. As a consequence, our model significantly outperforms all competitive baselines thanks to its lightweight architecture.

\begin{table*}[t]
\centering
\resizebox{0.85\linewidth}{!}{
\begin{tabular}{ccccccccccc}
\thickhline
\multirow{2}{*}{Models}                                             & \multicolumn{5}{c}{DailyDialog}                                                                                                  & \multicolumn{5}{c}{ConvAI2}                                                                                                       \\
\cmidrule(lr){2-6}\cmidrule(lr){7-11} 
                                                                    & \multicolumn{1}{c}{BLEU-1} & \multicolumn{1}{c}{BLEU-2} & \multicolumn{1}{c}{ROUGE-1} & \multicolumn{1}{c}{ROUGE-2} & ROUGE-L & \multicolumn{1}{c}{BLEU-1} & \multicolumn{1}{c}{BLEU-2} & \multicolumn{1}{c}{ROUGE-1} & \multicolumn{1}{c}{ROUGE-2} & ROUGE-L \\ \midrule
Full Model                                                          & \multicolumn{1}{c}{9.58}   & \multicolumn{1}{c}{3.40}    & \multicolumn{1}{c}{11.99}   & \multicolumn{1}{c}{2.50}     & 11.35   & \multicolumn{1}{c}{11.16}  & \multicolumn{1}{c}{4.62}   & \multicolumn{1}{c}{14.73}   & \multicolumn{1}{c}{3.25}    & 14.11   \\ \midrule
-$c$                                                       & \multicolumn{1}{c}{4.55}   & \multicolumn{1}{c}{1.62}   & \multicolumn{1}{c}{8.66}    & \multicolumn{1}{c}{2.08}    & 8.30     & \multicolumn{1}{c}{7.01}   & \multicolumn{1}{c}{2.65}   & \multicolumn{1}{c}{10.30}    & \multicolumn{1}{c}{2.58}    & 10.13   \\
-$z^S$                                                     & \multicolumn{1}{c}{6.35}   & \multicolumn{1}{c}{2.16}   & \multicolumn{1}{c}{10.73}   & \multicolumn{1}{c}{2.53}    & 10.27   & \multicolumn{1}{c}{7.11}   & \multicolumn{1}{c}{2.88}   & \multicolumn{1}{c}{11.30}    & \multicolumn{1}{c}{2.61}    & 10.80    \\
-$z^I$                                                     & \multicolumn{1}{c}{7.27}   & \multicolumn{1}{c}{2.50}    & \multicolumn{1}{c}{11.10}    & \multicolumn{1}{c}{2.51}    & 10.65   & \multicolumn{1}{c}{10.08}  & \multicolumn{1}{c}{3.95}   & \multicolumn{1}{c}{13.34}   & \multicolumn{1}{c}{2.97}    & 12.72   \\
-disentangle & \multicolumn{1}{c}{6.26}   & \multicolumn{1}{c}{2.19}   & \multicolumn{1}{c}{1.08}    & \multicolumn{1}{c}{2.41}    & 9.74    & \multicolumn{1}{c}{7.78}   & \multicolumn{1}{c}{3.06}   & \multicolumn{1}{c}{12.37}   & \multicolumn{1}{c}{2.77}    & 11.80    \\ \bottomrule
\end{tabular}
}
\vspace{-1mm}
\caption{Ablation study on DailyDialog and ConvAI2.}
\vspace{-1mm}
\label{tab:exp-abl}
\end{table*}

\subsection{Ablation Study}
To understand the impact of different variables on model performance and the effect of disentanglement, we compare the full model with the following variants:
(1) -$c$: the discrete latent variable is removed;
(2) -$z^S$: the context-sensitive latent variable is removed;
(3) -$z^I$: the context-independent latent variable is removed;
(4) -disentangle: the model is only trained with $\mathcal{L}_{ELBO}$.
All models are evaluated under the zero-resource setting to gain a full grasp of the transferability of our model. Table \ref{tab:exp-abl} reports the evaluation results. 
We can conclude that:
(1) The discrete latent variable $c$ plays a crucial role in both datasets, as eliminating the variable causes a dramatic performance drop. It is reasonable since our model can capture state transitions between utterances thanks to the latent structure.
(2) Though the removal of the context-sensitive or the context-independent variables both results in a performance drop, the context-sensitive latent variable $z^{S}$ is much more beneficial because it can eliminate context-independent information from the dialogue structure, allowing the model to be more transferable. 
(3) The self-supervised tasks designed for disentanglement are effective because removing them leads to a decline in performance.

\begin{table}[h!]
\centering
\resizebox{1\linewidth}{!}{
\begin{tabular}{ccccccc}
\toprule
\multirow{2}{*}{Models}                        & \multicolumn{3}{c}{DailyDialog}                                                                                                  & \multicolumn{3}{c}{ConvAI2}                                                                                                       \\ \cmidrule(lr){2-4}\cmidrule(lr){5-7}
                                               & \multicolumn{1}{c}{BLEU-1} & \multicolumn{1}{c}{BLEU-2} & ROUGE-L & \multicolumn{1}{c}{BLEU-1} & \multicolumn{1}{c}{BLEU-2} & ROUGE-L \\ \midrule
Full Model (w/o freeze)                         & \multicolumn{1}{c}{24.87}  & \multicolumn{1}{c}{16.70}  & 21.47   & \multicolumn{1}{c}{20.15}  & \multicolumn{1}{c}{9.41}   &  17.38   \\ 
Full Model (freeze)                        & \multicolumn{1}{c}{25.10}  & \multicolumn{1}{c}{16.18}  & 20.47   & \multicolumn{1}{c}{20.25}  & \multicolumn{1}{c}{9.57}   &  17.27   \\ 
-$z^S$ \& $z^I$ (w/o freeze)  & \multicolumn{1}{c}{18.07}  & \multicolumn{1}{c}{9.11}   &  15.98   & \multicolumn{1}{c}{18.05}  & \multicolumn{1}{c}{8.17}   &  16.28   \\ 
-$z^S$ \& $z^I$ (freeze) & \multicolumn{1}{c}{16.60}  & \multicolumn{1}{c}{7.19}   &  13.70   & \multicolumn{1}{c}{16.61}  & \multicolumn{1}{c}{8.08}   &  16.25   \\ 
\bottomrule
\end{tabular}
}
\caption{Evaluation results about the transferability on DailyDialog and ConvAI2.}
\label{tab:exp-transfer}
\end{table}

\subsection{Further Analysis on Transferability}
This part will move one step further to understand the transferability of the dialogue structure learned from the large-scale corpus. A dialogue structure with strong transferability is supposed to be well adapted to downstream tasks even without fine-tuning structure-related parameters, which is much more challenging. Therefore, to further verify the transferability of our methods, we freeze the parameters of $f_{c-trans}$ (in Eq.\ref{eq:discrete1}) and $f_{c-mlp}$ (in Eq.\ref{eq:discrete2}), and only fine-tune the remaining parameters on downstream tasks. We additionally provide a variant in which the context-sensitive and context-independent latent variables are removed. Table \ref{tab:exp-transfer} reports the results. It can be seen that our model still performs well once some parameters are frozen, thanks to the proposed disentangled structure learning framework. On the other hand, when context-sensitive and context-independent latent variables are removed, freezing the parameters will result in a significant performance drop. The reason is that our proposed approach can disentangle the holistic information from the dialogue structure, allowing it to be used for various downstream tasks.

\begin{figure}
\centering
\includegraphics[width=1.0\linewidth]{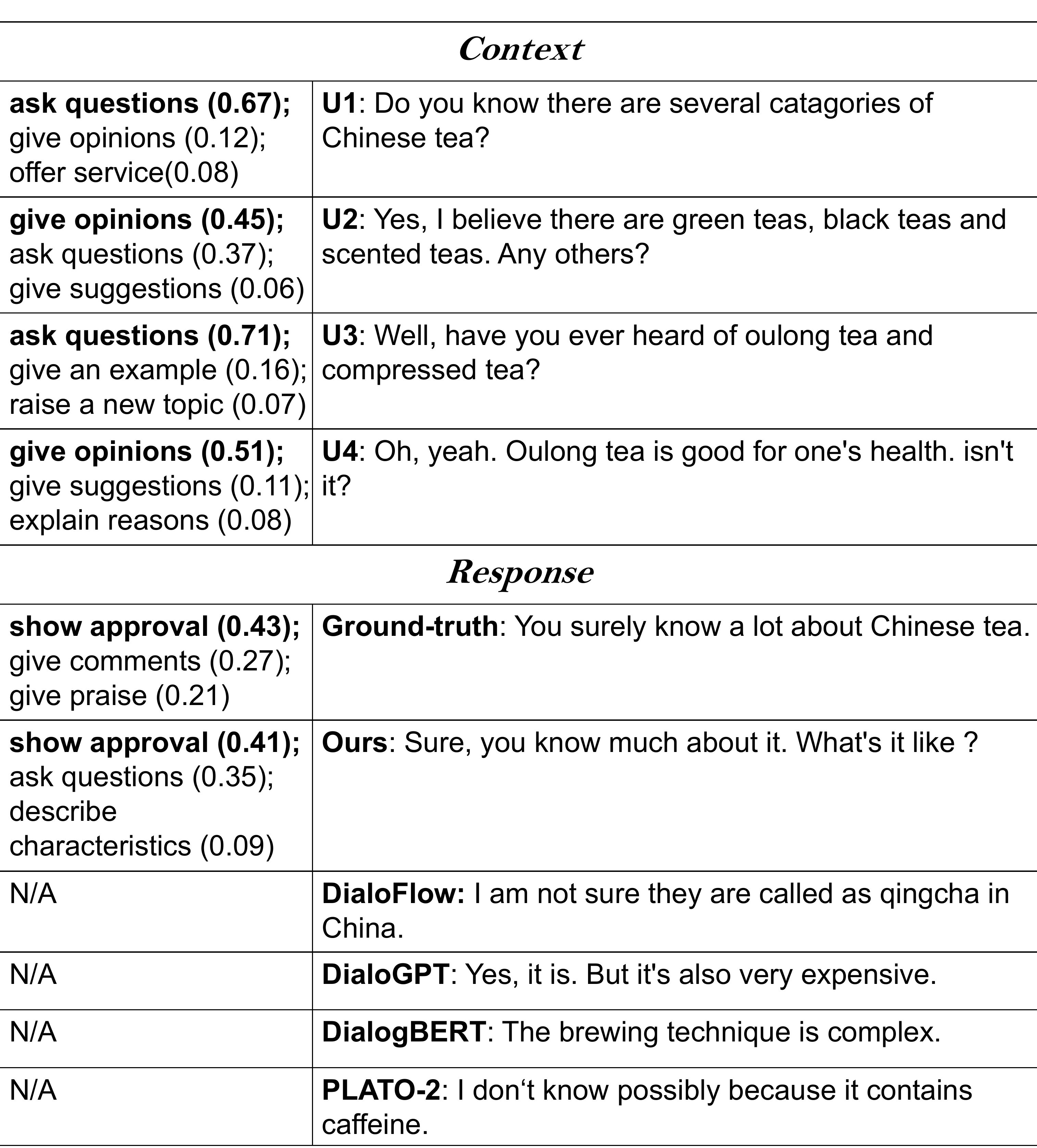}
\caption{A case from the Test set of DailyDialog.}
\label{fig:case-study}
\end{figure}

\subsection{Case Study on Iterpretability}
\label{sec:case}
Figure \ref{fig:case-study} shows an example from the test set of DailyDialog.
The left column gives the states of the discrete latent variable along with the probabilities predicted by our model and the right column shows the corresponding utterances. We employ human experts to consistently interpret each state by going through utterances assigned to the same state. We only present the top-$3$ states due to space constraints. We can observe that: 
(1) our model can accurately infer the states of context utterances and the ground-truth response given the posterior distribution of $c$; 
(2) thanks to the dialogue structure our model can give a more appropriate response to catch up with the context than baseline models.
\section{Conclusion}

We propose a novel dialogue model with structural bias which is explainable to humans and easily transferable to general dialogue tasks. Empirical experiments on two benchmark datasets indicate that our model with only 22\% parameters outperforms the strongest baseline DialoFlow in both decoding speed and response quality measured by automatic and human evaluations. We further show that the learned latent structure enjoys superior transferability and interpretability compared to the conventional methods.

\section*{Limitations}

In this paper, we propose a dialogue pre-training model that featured a discrete transition structure. By introducing a series of latent variables into the pre-training process, the pre-trained model could be easily adapted into downstream application scenarios in a transparent and interpretable way. However, all technologies built upon the large-scale PLM more or less inherit their potential harms~\cite{bender2021dangers}. Besides, we identify some limitations within our work and describe them below:

(1) Although the conversation flow is discrete and interpretable, it is laborious to interpret the implication of each state in the conversation flow by going through utterances assigned to the same state. Besides, large-scale human evaluation in our experiments is also costly and time-consuming.

(2) The vocabulary size of the latent conversation flow $N$ is an important parameter that requires to be carefully tuned, especially when the model is agnostic to the downstream task. The optimum $N$ may vary according to the different languages or different domains. It is particularly challenging to shift to uncommon languages because of its reliance on large-scale pre-training corpus.

(3) In this paper, we use discrete latent variables to model the conversation flow for better interpretability. From another perspective, we are distributing utterances into clusters according to their latent flow variable $c$. However, this may incur high intra-cluster diversity, especially when generalizing to out-of-distribution data. A possible remedy is to introduce some regularity in training $f_{c-mlp}$ to push its predicted distribution towards one-hot distribution.

(4) The adoption of our method can lead to better dialogue systems that improve the quality of life for many people. But our method could also affect the human interlocutors in a negative way if used for malicious intent. We advise that any plan to apply our method should consider carefully all potential groups of stakeholders as well as the risk profiles of applied domains to maximize the overall positive impacts.

\section*{Ethics Statement}
This paper studies open-domain dialogue pre-training and proposes a disentangled structure learning framework that allows the transformer architecture to capture the prior knowledge about state transition in a large-scale dialogue corpus. There are no ethical issues with this research. The datasets we used are commonly utilized by other researchers and are typically accessible to the public. The proposed approach does not introduce ethical or societal prejudice.

\section*{Acknowledgements}
We appreciate the anonymous reviewers for their constructive comments. This work was supported by National Natural Science Foundation of China~(NSFC Grant No. 62122089 and No. 61876196), Beijing Outstanding Young Scientist Program NO. BJJWZYJH012019100020098, and Intelligent Social Governance Platform, Major Innovation \& Planning Interdisciplinary Platform for the ``Double-First Class'' Initiative, Renmin University of China. This work was also supported in part by Independent Research Fund Denmark under agreement 8048-00038B. This work was also supported by the National Key Research and Development Program of China~(No. 2020AAA0106600). We wish to acknowledge the support provided and contribution made by Public Policy and Decision-making Research Lab of RUC. Rui Yan is supported by Beijing Academy of Artificial Intelligence (BAAI) and Tencent Collaborative Research Fund.

\bibliography{anthology,custom}

\begin{thebibliography}{45}
\expandafter\ifx\csname natexlab\endcsname\relax\def\natexlab#1{#1}\fi

\bibitem[{Adiwardana et~al.(2020)Adiwardana, Luong, So, Hall, Fiedel,
  Thoppilan, Yang, Kulshreshtha, Nemade, Lu et~al.}]{adiwardana2020towards}
Daniel Adiwardana, Minh-Thang Luong, David~R So, Jamie Hall, Noah Fiedel, Romal
  Thoppilan, Zi~Yang, Apoorv Kulshreshtha, Gaurav Nemade, Yifeng Lu, et~al.
  2020.
\newblock Towards a human-like open-domain chatbot.
\newblock \emph{arXiv preprint arXiv:2001.09977}.

\bibitem[{Bao et~al.(2020)Bao, He, Wang, Wu, and Wang}]{bao2019plato}
Siqi Bao, Huang He, Fan Wang, Hua Wu, and Haifeng Wang. 2020.
\newblock \href {https://doi.org/10.18653/v1/2020.acl-main.9} {{PLATO}:
  Pre-trained dialogue generation model with discrete latent variable}.
\newblock In \emph{Proceedings of the 58th Annual Meeting of the Association
  for Computational Linguistics}, pages 85--96, Online. Association for
  Computational Linguistics.

\bibitem[{Bao et~al.(2021)Bao, He, Wang, Wu, Wang, Wu, Guo, Liu, and
  Xu}]{bao2021plato2}
Siqi Bao, Huang He, Fan Wang, Hua Wu, Haifeng Wang, Wenquan Wu, Zhen Guo,
  Zhibin Liu, and Xinchao Xu. 2021.
\newblock \href {https://doi.org/10.18653/v1/2021.findings-acl.222} {{PLATO-2}:
  Towards building an open-domain chatbot via curriculum learning}.
\newblock In \emph{Findings of the Association for Computational Linguistics:
  ACL-IJCNLP 2021}, pages 2513--2525, Online. Association for Computational
  Linguistics.

\bibitem[{Bender et~al.(2021)Bender, Gebru, McMillan-Major, and
  Shmitchell}]{bender2021dangers}
Emily~M Bender, Timnit Gebru, Angelina McMillan-Major, and Shmargaret
  Shmitchell. 2021.
\newblock On the dangers of stochastic parrots: Can language models be too big?
\newblock In \emph{Proceedings of the 2021 ACM Conference on Fairness,
  Accountability, and Transparency}, pages 610--623.

\bibitem[{Chen et~al.(2019)Chen, Li, Grosse, and Duvenaud}]{chen2019isolating}
Ricky~TQ Chen, Xuechen Li, Roger Grosse, and David Duvenaud. 2019.
\newblock Isolating sources of disentanglement in vaes.
\newblock In \emph{Proceedings of the 32nd International Conference on Neural
  Information Processing Systems}, pages 2615--2625.

\bibitem[{Dinan et~al.(2020)Dinan, Logacheva, Malykh, Miller, Shuster, Urbanek,
  Kiela, Szlam, Serban, Lowe et~al.}]{dinan2020second}
Emily Dinan, Varvara Logacheva, Valentin Malykh, Alexander Miller, Kurt
  Shuster, Jack Urbanek, Douwe Kiela, Arthur Szlam, Iulian Serban, Ryan Lowe,
  et~al. 2020.
\newblock The second conversational intelligence challenge (convai2).
\newblock In \emph{The NeurIPS'18 Competition}, pages 187--208. Springer.

\bibitem[{Dinan et~al.(2019)Dinan, Roller, Shuster, Fan, Auli, and
  Weston}]{dinan2018wizard}
Emily Dinan, Stephen Roller, Kurt Shuster, Angela Fan, Michael Auli, and Jason
  Weston. 2019.
\newblock Wizard of wikipedia: Knowledge-powered conversational agents.
\newblock In \emph{ICLR}.

\bibitem[{Fleiss(1971)}]{fleiss1971measuring}
Joseph~L Fleiss. 1971.
\newblock Measuring nominal scale agreement among many raters.
\newblock \emph{Psychological bulletin}, 76(5):378.

\bibitem[{Gehring et~al.(2017)Gehring, Auli, Grangier, Yarats, and
  Dauphin}]{gehring2017convolutional}
Jonas Gehring, Michael Auli, David Grangier, Denis Yarats, and Yann~N Dauphin.
  2017.
\newblock Convolutional sequence to sequence learning.
\newblock In \emph{Proceedings of the 34th International Conference on Machine
  Learning-Volume 70}, pages 1243--1252. JMLR. org.

\bibitem[{Gu et~al.(2021)Gu, Yoo, and Ha}]{gu2021dialogbert}
Xiaodong Gu, Kang~Min Yoo, and Jung-Woo Ha. 2021.
\newblock Dialog{BERT}: Discourse-aware response generation via learning to
  recover and rank utterances.

\bibitem[{Hoffman et~al.(2013)Hoffman, Blei, Wang, and
  Paisley}]{hoffman2013stochastic}
Matthew~D Hoffman, David~M Blei, Chong Wang, and John Paisley. 2013.
\newblock Stochastic variational inference.
\newblock \emph{Journal of Machine Learning Research}, 14(5).

\bibitem[{Jang et~al.(2017)Jang, Gu, and Poole}]{jang2017categorical}
Eric Jang, Shixiang Gu, and Ben Poole. 2017.
\newblock \href {https://openreview.net/forum?id=rkE3y85ee} {Categorical
  reparameterization with gumbel-softmax}.
\newblock In \emph{International Conference on Learning Representations}.

\bibitem[{Kingma and Ba(2015)}]{kingma2014adam}
Diederik~P Kingma and Jimmy Ba. 2015.
\newblock Adam: A method for stochastic optimization.
\newblock In \emph{ICLR}.

\bibitem[{Kingma and Welling(2013)}]{kingma2013auto}
Diederik~P Kingma and Max Welling. 2013.
\newblock Auto-encoding variational bayes.
\newblock \emph{arXiv preprint arXiv:1312.6114}.

\bibitem[{Lavie and Agarwal(2007)}]{lavie2007meteor}
Alon Lavie and Abhaya Agarwal. 2007.
\newblock Meteor: An automatic metric for mt evaluation with high levels of
  correlation with human judgments.
\newblock In \emph{Proceedings of the second workshop on statistical machine
  translation}, pages 228--231.

\bibitem[{Li et~al.(2015)Li, Galley, Brockett, Gao, and
  Dolan}]{li2015diversity}
Jiwei Li, Michel Galley, Chris Brockett, Jianfeng Gao, and Bill Dolan. 2015.
\newblock A diversity-promoting objective function for neural conversation
  models.
\newblock \emph{NAACL}, pages 110--119.

\bibitem[{Li et~al.(2017)Li, Su, Shen, Li, Cao, and Niu}]{li2017dailydialog}
Yanran Li, Hui Su, Xiaoyu Shen, Wenjie Li, Ziqiang Cao, and Shuzi Niu. 2017.
\newblock Dailydialog: A manually labelled multi-turn dialogue dataset.
\newblock In \emph{Proceedings of the Eighth International Joint Conference on
  Natural Language Processing (Volume 1: Long Papers)}, pages 986--995.

\bibitem[{Li et~al.(2021)Li, Zhang, Fei, Feng, and Zhou}]{li2021conversations}
Zekang Li, Jinchao Zhang, Zhengcong Fei, Yang Feng, and Jie Zhou. 2021.
\newblock \href {https://doi.org/10.18653/v1/2021.acl-long.11} {Conversations
  are not flat: Modeling the dynamic information flow across dialogue
  utterances}.
\newblock In \emph{Proceedings of the 59th Annual Meeting of the Association
  for Computational Linguistics and the 11th International Joint Conference on
  Natural Language Processing (Volume 1: Long Papers)}, pages 128--138, Online.
  Association for Computational Linguistics.

\bibitem[{Lin(2004)}]{lin2004rouge}
Chin-Yew Lin. 2004.
\newblock Rouge: A package for automatic evaluation of summaries.
\newblock In \emph{Text summarization branches out}, pages 74--81.

\bibitem[{Papineni et~al.(2002)Papineni, Roukos, Ward, and
  Zhu}]{papineni2002bleu}
Kishore Papineni, Salim Roukos, Todd Ward, and Wei-Jing Zhu. 2002.
\newblock Bleu: a method for automatic evaluation of machine translation.
\newblock In \emph{Proceedings of the 40th annual meeting on association for
  computational linguistics}, pages 311--318. Association for Computational
  Linguistics.

\bibitem[{Qiu et~al.(2020)Qiu, Zhao, Shi, Liang, Shi, Yuan, Yu, and
  Zhu}]{qiu2020structured}
Liang Qiu, Yizhou Zhao, Weiyan Shi, Yuan Liang, Feng Shi, Tao Yuan, Zhou Yu,
  and Song-chun Zhu. 2020.
\newblock Structured attention for unsupervised dialogue structure induction.
\newblock In \emph{Proceedings of the 2020 Conference on Empirical Methods in
  Natural Language Processing (EMNLP)}, pages 1889--1899.

\bibitem[{Radford et~al.(2019)Radford, Wu, Child, Luan, Amodei, and
  Sutskever}]{radford2019language}
Alec Radford, Jeffrey Wu, Rewon Child, David Luan, Dario Amodei, and Ilya
  Sutskever. 2019.
\newblock Language models are unsupervised multitask learners.

\bibitem[{Ritter et~al.(2011)Ritter, Cherry, and Dolan}]{ritter2011data}
Alan Ritter, Colin Cherry, and William~B Dolan. 2011.
\newblock Data-driven response generation in social media.
\newblock In \emph{Proceedings of the 2011 Conference on Empirical Methods in
  Natural Language Processing}, pages 583--593.

\bibitem[{See et~al.(2019)See, Roller, Kiela, and Weston}]{see2019makes}
Abigail See, Stephen Roller, Douwe Kiela, and Jason Weston. 2019.
\newblock \href {https://doi.org/10.18653/v1/N19-1170} {What makes a good
  conversation? how controllable attributes affect human judgments}.
\newblock In \emph{Proceedings of the 2019 Conference of the North {A}merican
  Chapter of the Association for Computational Linguistics: Human Language
  Technologies, Volume 1 (Long and Short Papers)}, pages 1702--1723,
  Minneapolis, Minnesota. Association for Computational Linguistics.

\bibitem[{Serban et~al.(2016)Serban, Sordoni, Bengio, Courville, and
  Pineau}]{serban2016building}
Iulian~Vlad Serban, Alessandro Sordoni, Yoshua Bengio, Aaron~C Courville, and
  Joelle Pineau. 2016.
\newblock Building end-to-end dialogue systems using generative hierarchical
  neural network models.
\newblock In \emph{AAAI}, volume~16, pages 3776--3784.

\bibitem[{Serban et~al.(2017)Serban, Sordoni, Lowe, Charlin, Pineau, Courville,
  and Bengio}]{serban2017hierarchical}
Iulian~Vlad Serban, Alessandro Sordoni, Ryan Lowe, Laurent Charlin, Joelle
  Pineau, Aaron~C Courville, and Yoshua Bengio. 2017.
\newblock A hierarchical latent variable encoder-decoder model for generating
  dialogues.
\newblock In \emph{AAAI}, pages 3295--3301.

\bibitem[{Shang et~al.(2015)Shang, Lu, and Li}]{shangL2015neural}
Lifeng Shang, Zhengdong Lu, and Hang Li. 2015.
\newblock Neural responding machine for short-text conversation.
\newblock In \emph{ACL}, pages 1577--1586.

\bibitem[{Shi et~al.(2019)Shi, Zhao, and Yu}]{shi2019unsupervised}
Weiyan Shi, Tiancheng Zhao, and Zhou Yu. 2019.
\newblock Unsupervised dialog structure learning.
\newblock In \emph{Proceedings of the 2019 Conference of the North American
  Chapter of the Association for Computational Linguistics: Human Language
  Technologies, Volume 1 (Long and Short Papers)}, pages 1797--1807.

\bibitem[{Sun et~al.(2021)Sun, Shan, Tang, Hu, Dai, Yu, Sun, Huang, and
  Si}]{sun2021unsupervised}
Yajing Sun, Yong Shan, Chengguang Tang, Yue Hu, Yinpei Dai, Jing Yu, Jian Sun,
  Fei Huang, and Luo Si. 2021.
\newblock Unsupervised learning of deterministic dialogue structure with
  edge-enhanced graph auto-encoder.
\newblock In \emph{Proceedings of the Thirty-Fifth Conference on Association
  for the Advancement of Artificial Intelligence (AAAI)}, pages 13869--13877.

\bibitem[{Sutskever et~al.(2014)Sutskever, Vinyals, and
  Le}]{sutskever2014sequence}
Ilya Sutskever, Oriol Vinyals, and Quoc~V Le. 2014.
\newblock Sequence to sequence learning with neural networks.
\newblock In \emph{Advances in neural information processing systems}, pages
  3104--3112.

\bibitem[{Tao et~al.(2018)Tao, Gao, Shang, Wu, Zhao, and Yan}]{tao2018get}
Chongyang Tao, Shen Gao, Mingyue Shang, Wei Wu, Dongyan Zhao, and Rui Yan.
  2018.
\newblock Get the point of my utterance! learning towards effective responses
  with multi-head attention mechanism.
\newblock In \emph{IJCAI}, pages 4418--4424.

\bibitem[{Vaswani et~al.(2017)Vaswani, Shazeer, Parmar, Uszkoreit, Jones,
  Gomez, Kaiser, and Polosukhin}]{vaswani2017attention}
Ashish Vaswani, Noam Shazeer, Niki Parmar, Jakob Uszkoreit, Llion Jones,
  Aidan~N Gomez, {\L}ukasz Kaiser, and Illia Polosukhin. 2017.
\newblock Attention is all you need.
\newblock In \emph{NIPS}, pages 5998--6008.

\bibitem[{Vinyals and Le(2015)}]{vinyals2015neural}
Oriol Vinyals and Quoc Le. 2015.
\newblock A neural conversational model.
\newblock \emph{arXiv preprint arXiv:1506.05869}.

\bibitem[{Wang et~al.(2018)Wang, Liu, Huang, and Nie}]{wang2018learning}
Yansen Wang, Chenyi Liu, Minlie Huang, and Liqiang Nie. 2018.
\newblock Learning to ask questions in open-domain conversational systems with
  typed decoders.
\newblock In \emph{Proceedings of the 56th Annual Meeting of the Association
  for Computational Linguistics (Volume 1: Long Papers)}, pages 2193--2203.

\bibitem[{Wu et~al.(2020)Wu, Wu, Qi, Cui, and Huang}]{wu2020attentive}
Chuhan Wu, Fangzhao Wu, Tao Qi, Xiaohui Cui, and Yongfeng Huang. 2020.
\newblock Attentive pooling with learnable norms for text representation.
\newblock In \emph{Proceedings of the 58th Annual Meeting of the Association
  for Computational Linguistics}, pages 2961--2970.

\bibitem[{Xu et~al.(2019)Xu, Wu, Tao, Hu, Schuerman, and Wang}]{xu2019neural}
Can Xu, Wei Wu, Chongyang Tao, Huang Hu, Matt Schuerman, and Ying Wang. 2019.
\newblock Neural response generation with meta-words.
\newblock \emph{arXiv preprint arXiv:1906.06050}.

\bibitem[{Xu et~al.(2021)Xu, Lei, Wang, Niu, Wu, and Che}]{xu2021discovering}
Jun Xu, Zeyang Lei, Haifeng Wang, Zheng-Yu Niu, Hua Wu, and Wanxiang Che. 2021.
\newblock Discovering dialog structure graph for coherent dialog generation.
\newblock In \emph{Proceedings of the 59th Annual Meeting of the Association
  for Computational Linguistics and the 11th International Joint Conference on
  Natural Language Processing (Volume 1: Long Papers)}, pages 1726--1739.

\bibitem[{Zhang et~al.(2019)Zhang, Lan, Pang, Guo, and Cheng}]{zhang2019recosa}
Hainan Zhang, Yanyan Lan, Liang Pang, Jiafeng Guo, and Xueqi Cheng. 2019.
\newblock Recosa: Detecting the relevant contexts with self-attention for
  multi-turn dialogue generation.
\newblock In \emph{Proceedings of the 57th Annual Meeting of the Association
  for Computational Linguistics}, pages 3721--3730.

\bibitem[{Zhang et~al.(2018)Zhang, Guo, Fan, Lan, Xu, and
  Cheng}]{zhang2018learning}
Ruqing Zhang, Jiafeng Guo, Yixing Fan, Yanyan Lan, Jun Xu, and Xueqi Cheng.
  2018.
\newblock Learning to control the specificity in neural response generation.
\newblock In \emph{Proceedings of the 56th Annual Meeting of the Association
  for Computational Linguistics (Volume 1: Long Papers)}, pages 1108--1117.

\bibitem[{Zhang et~al.(2020)Zhang, Sun, Galley, Chen, Brockett, Gao, Gao, Liu,
  and Dolan}]{zhang2020dialogpt}
Yizhe Zhang, Siqi Sun, Michel Galley, Yen-Chun Chen, Chris Brockett, Xiang Gao,
  Jianfeng Gao, Jingjing Liu, and Bill Dolan. 2020.
\newblock \href {https://doi.org/10.18653/v1/2020.acl-demos.30} {{DIALOGPT} :
  Large-scale generative pre-training for conversational response generation}.
\newblock In \emph{Proceedings of the 58th Annual Meeting of the Association
  for Computational Linguistics: System Demonstrations}, pages 270--278,
  Online. Association for Computational Linguistics.

\bibitem[{Zhao et~al.(2017)Zhao, Zhao, and Eskenazi}]{zhao2017learning}
Tiancheng Zhao, Ran Zhao, and Maxine Eskenazi. 2017.
\newblock Learning discourse-level diversity for neural dialog models using
  conditional variational autoencoders.
\newblock In \emph{ACL}, pages 654--664.

\bibitem[{Zhao et~al.(2020{\natexlab{a}})Zhao, Wu, Tao, Xu, Zhao, and
  Yan}]{zhao2019low}
Xueliang Zhao, Wei Wu, Chongyang Tao, Can Xu, Dongyan Zhao, and Rui Yan.
  2020{\natexlab{a}}.
\newblock Low-resource knowledge-grounded dialogue generation.
\newblock In \emph{International Conference on Learning Representations}.

\bibitem[{Zhao et~al.(2020{\natexlab{b}})Zhao, Wu, Xu, Tao, Zhao, and
  Yan}]{zhao2020knowledge}
Xueliang Zhao, Wei Wu, Can Xu, Chongyang Tao, Dongyan Zhao, and Rui Yan.
  2020{\natexlab{b}}.
\newblock Knowledge-grounded dialogue generation with pre-trained language
  models.
\newblock In \emph{Proceedings of the 2020 Conference on Empirical Methods in
  Natural Language Processing (EMNLP)}, pages 3377--3390.

\bibitem[{Zhou et~al.(2017)Zhou, Huang, Zhang, Zhu, and
  Liu}]{zhou2017emotional}
Hao Zhou, Minlie Huang, Tianyang Zhang, Xiaoyan Zhu, and Bing Liu. 2017.
\newblock Emotional chatting machine: Emotional conversation generation with
  internal and external memory.
\newblock \emph{arXiv preprint arXiv:1704.01074}.

\bibitem[{Zhu et~al.(2020)Zhu, Min, Kadav, and Graf}]{zhu2020s3vae}
Yizhe Zhu, Martin~Renqiang Min, Asim Kadav, and Hans~Peter Graf. 2020.
\newblock S3vae: Self-supervised sequential vae for representation
  disentanglement and data generation.
\newblock In \emph{Proceedings of the IEEE/CVF Conference on Computer Vision
  and Pattern Recognition}, pages 6538--6547.

\end{thebibliography}
\bibliographystyle{acl_natbib}

\clearpage
\appendix

\section{Details of Datasets}
\label{sec:app-data}

We follow~\citet{zhang2020dialogpt} to adopt the Reddit comments as our pre-training data, which contains various domains and topics. 
We crawl the online discussions over a period spanning from $2011$ through $2016$,
and there are $60,579,645$ and $685,881$ dialogues in the training set and the validation set respectively. 
Each dialogue has $7.9$ utterances on average, with each utterance including $27.5$ words.

We evaluate our model on two benchmark datasets for multi-turn dialogue generation.
(1) \textbf{DailyDialog Dataset.} This dataset is manually labeled and contains conversations about daily life~\cite{li2017dailydialog}. 
This dataset is split into training set, validation set, and test set by the data owners. 
(2) \textbf{ConvAI2 Dataset.} This dataset is collected by having two workers at Amazon Mechanical Turk chat with each other based on their assigned profiles~\cite{dinan2020second}. The profiles define speakers’ personas and provide characteristic knowledge for dialogues. 
Since the test set of ConvAI2 has not been made public, we randomly select $5\%$ sessions from the original training set as our validation set and use the original validation set as our test set. 

To facilitate reproducibility, we adopt the datasets shared at ParlAI\footnote{\url{https://github.com/facebookresearch/ParlAI/tree/main/parlai/tasks}} and conduct pre-processing with the code available there. More statistics of the two datasets are shown in Table \ref{tab:stat}.

\begin{table}[h!]
\centering
\resizebox{1\linewidth}{!}{
\begin{tabular}{lcccccc}
\toprule
\multicolumn{1}{c}{\multirow{2}{*}{Statistics}} & \multicolumn{3}{c}{DailyDialog}                                 & \multicolumn{3}{c}{ConvAI2}                                        \\ \cmidrule(lr){2-4}\cmidrule(lr){5-7} 
\multicolumn{1}{c}{}                            & \multicolumn{1}{c}{Train}  & \multicolumn{1}{c}{Val}   & Test  & \multicolumn{1}{c}{Train}   & \multicolumn{1}{c}{Val}   & Test  \\ 
\midrule
\# Sessions                                      & \multicolumn{1}{c}{22,236} & \multicolumn{1}{c}{2,000} & 2,000 & \multicolumn{1}{c}{16,985}  & \multicolumn{1}{c}{893}   & 1,000 \\ \midrule
\# Turns                                         & \multicolumn{1}{c}{87,170} & \multicolumn{1}{c}{8,069} & 7,740 & \multicolumn{1}{c}{124,877} & \multicolumn{1}{c}{6,561} & 7,801 \\ \midrule
\# Turns / Session                               & \multicolumn{1}{c}{3.9}    & \multicolumn{1}{c}{4.0}   & 3.9   & \multicolumn{1}{c}{7.4}     & \multicolumn{1}{c}{7.3}   & 7.8   \\ \midrule
\# Words / Turn                                  & \multicolumn{1}{c}{13.0}   & \multicolumn{1}{c}{12.9}  & 13.1  & \multicolumn{1}{c}{11.1}    & \multicolumn{1}{c}{11.0}  & 11.7  \\
\bottomrule
\end{tabular}
}
\caption{Statistics of the two datasets.}
\label{tab:stat}
\end{table}

\section{More Implementation Details}
\label{sec:app-imp}

The total number of discrete latent states (i.e., $N$) is set as $100$ for all experiments. The dimension of context-sensitive and context-independent latent variables are both set as $768$. The embedding size of discrete latent states is set as $768$. The MLP network in $f_{c-mlp}$ has two layers with the input, hidden and output dimensions being $768$, $100$ and $100$ respectively. We choose GPT-$2$ (117M) as the backbone of our model. All models are learned with Adam~\cite{kingma2014adam} optimizer with $\beta_1=0.9$ and $\beta_2 = 0.999$. We set the initial temperature, the minimum temperature, and the anneal rate of Gumbel Softmax as $1.0$, $0.5$, and $4e-5$ respectively. In the training phase, the batch size is set as $64$, and the learning rate is set as $2e-5$. In the test phase, we employ beam search in response decoding with beam size $=5$. Early stopping on validation is adopted as a regularization strategy. In our experiments, the profiles in ConvAI2 are concatenated as a long sequence and serve as the first sentence in a session. In both DailyDialog and ConvAI2, $7$ turns before an utterance are used as conversation history. All utterances are padded to a maximum length of $32$ tokens.

\section{Derivation of ELBO}
\label{sec:app-elbo}

\begin{equation}
\begin{aligned}
\mathcal{L}_{ELBO} = &\sum_{t=1}^{n}\mathbb{E}_{q(z^I_{t},z^{S})} \log p(u_t | u_{<t}, z^{I}_{t},z^{S}) \\
&- D_{\mathrm{KL}}(q(c,z^{I}, z^{S}) \| p(c,z^{I},z^{S})).
\end{aligned}
\end{equation}

According to the mean-field approximation, $q(c,z^{I}, z^{S}) \sim q(c)q(z^{I})q(z^S)$. Therefore, the last term can be re-written as:

\begin{equation}
\scriptsize
\begin{aligned}
& D_{\mathrm{KL}}(q(c,z^{I}, z^{S}) \| p(c,z^{I},z^{S})) \\
&=\sum q(c|X) \int q(z^{I}|X) q(z^{s}|X) \log \frac{q(c|X) q(z^{I}|X) q(z^{s}|X)}{p(c) p(z^{I}) p(z^{s})} \Delta z^{I} \Delta z^{s} \\
&=\sum q(c|X) \log \frac{q(c|X)}{p(c)}+\sum q(c|X) \int q(z^{I}|X) \log \frac{q(z^{I}|X)}{p(z^{I})} \Delta z^{I} \\
& +\int q(z^{S}|X) \log \frac{q(z^{S}|X)}{p(z^{S})} \Delta z^{S} \\
&=\sum_{t=1}^{n} D_{K L}(q(c_{t}|X) \| p(c_{t}|c_{<t}))+D_{K L}(q(z^{s} | X) \| p(z^{s})) \\
& +\sum_{t=1}^{n} \mathbb{E}_{q(c_{t})} D_{K L}(q(z_{t}^{I} | X) \| p(z_{t}^{I} | u_{<t}, c_{t}))
\end{aligned}
\end{equation}

\end{document}